\documentclass[conference]{IEEEtran}
\IEEEoverridecommandlockouts
\usepackage{cite}
\usepackage{amsmath,amssymb,amsfonts}
\usepackage{algorithmic}
\usepackage{graphicx}
\usepackage{textcomp}
\usepackage{multirow}
\usepackage{bm}

\usepackage{xcolor}
\def\BibTeX{{\rm B\kern-.05em{\sc i\kern-.025em b}\kern-.08em
    T\kern-.1667em\lower.7ex\hbox{E}\kern-.125emX}}

\begin{document}

\title{A Question Answering Based Pipeline for Comprehensive Chinese EHR Information Extraction}

\author{\IEEEauthorblockN{Huaiyuan Ying}
\IEEEauthorblockA{\textit{Center for Statistical Science} \\
\textit{Tsinghua University}\\
Beijing, China \\
yinghy22@mails.tsinghua.edu.cn}
\and
\IEEEauthorblockN{Sheng Yu}
\IEEEauthorblockA{\textit{Center for Statistical Science} \\
\textit{Tsinghua University}\\
Beijing, China \\
syu@tsinghua.edu.cn}
}

\maketitle

\begin{abstract}
Electronic health records (EHRs) hold significant value for research and applications. As a new way of information extraction, question answering (QA) can extract more flexible information than conventional methods and is more accessible to clinical researchers, but its progress is impeded by the scarcity of annotated data. In this paper, we propose a novel approach that automatically generates training data for transfer learning of QA models. Our pipeline incorporates a preprocessing module to handle challenges posed by extraction types that is not readily compatible with extractive QA frameworks, including cases with discontinuous answers and many-to-one relationships. The obtained QA model exhibits excellent performance on subtasks of information extraction in EHRs, and it can effectively handle few-shot or zero-shot settings involving yes-no questions. Case studies and ablation studies demonstrate the necessity of each component in our design, and the resulting model is deemed suitable for practical use.
\end{abstract}


\section{Introduction}

Information extraction (IE) from electronic health records (EHRs) aims to convert free text content to structured data to facilitate diverse downstream tasks and analyses in healthcare research and services, such as developing patient data registries and EHR-linked biobanks. Conventionally, IE encompasses basic tasks such as named entity recognition (NER), entity linking, coreference resolution, and relation extraction (RE) \cite{singh2018natural} to extract entities, entity types, relations, events, and other valuable information. However, many biomedical studies require information that is significantly more sophisticated than what the above tasks can extract\cite{ouyang2022overview,nicholson2020constructing}. Combining basic tasks to extract more sophisticated information not only requires careful designs but is also difficult for users without a background in natural language processing (NLP). With the advancement of pretrained language models, question answering (QA) models can achieve impressive performance in extracting information that are more flexible and more complex than conventional tasks\cite{ishwari2019advances,athenikos2010biomedical}. It also has the advantage of being capable of accepting natural language queries, which can significantly lower the barrier to using IE in EHRs. Notably, the advent of large language models (LLMs) further enhanced the ability to summarize information \cite{brown2020language,chung2022scaling,touvron2023llama,xu2023wizardlm}, opening new possibilities in IE for EHRs.

QA tasks generally assume a data format that includes a context, a question, and an answer. Depending on how the answer is obtained, QA models can be classified as extractive or generative, where extractive models must identify the answer as a substring from the context, and generative models generate the answer autoregressively conditioning on the context and the question\cite{jin2022biomedical}. As a result, extractive QA is not as flexible as generative QA, but is also less prone to hallucination, which refers to generative models' behavior of generating fake answers based on its model parameters. As the goal of IE in EHRs is often extracting answers from the EHR text, we focus on extractive QA in this work to avoid potential hallucinations.

Training QA models requires the above-mentioned data format: each sample is comprised of a context, a question, and an answer. As QA models aim to answer a wide range of questions given in natural language, their training samples require more diversity than the other IE tasks\cite{rogers2023qa}. As a result, annotating QA data requires significantly more thought and takes more time than the basic IE tasks, making the development of QA models very costly. Fine-tuning LLMs for the biomedical domain also requires QA data, as LLMs like ChatGPT are obtained by fine-tuning unsupervised base models with QA data. Although LLMs are much more general-purpose than previous QA models and exhibit surprisingly good performance on medical data out-of-the-box, they still need to be further fine-tuned with QA data on EHRs to deliver satisfactory results.

In this paper, we propose a pipeline to automatically generate QA data from EHRs, and apply transfer learning on pretrained general domain QA models for IE tasks in EHRs. We leverage the fact that many projects have annotated abundant EHRs for basic IE tasks including NER and RE, and various NLP tools can perform these annotations as well. Based on the semantic types of the annotated entities, we design templates to convert entity and relation annotations to questions and answers, while we also prepare samples to train the model to judge answerability. Additionally, extractive QA models by default only identify a continuous span as the answer, but the EHR information to be extracted is usually discretely distributed. To address this issue, we incorporate additional preprocessing and post-processing in our pipeline. In brief, we break down paragraphs into sentences, exclude sentences without potential answers, and then merge the remaining answer spans. With transfer learning, we observed satisfactory performance. The final model demonstrates competency across various IE subtasks in EHRs and exhibits impressive generalization abilities. We summarize our contributions as follows:

\begin{itemize}
\item We propose a pipeline that utilizes accessible data annotations to train an extractive QA system, enabling it to assist in EHR IE. The pipeline accommodates diverse types of questions and performs multiple IE subtasks simultaneously.
\item We introduce processing techniques that equip the extractive QA system with the ability to judge answerability and extract discontinuous spans, thereby achieving superior performance compared to traditional IE methods.
\end{itemize}

\section{Related Work}

\subsection{EHR Information Extraction}

The extraction of information from health records and clinical documents has a long history. In its early stages, rule-based and expert-based systems were commonly used methods for Information Extraction (IE) \cite{sohn2009mayo,savova2010discovering}. Software such as cTAKES \cite{savova2010mayo} and MetaMap \cite{aronson2010overview}, equipped with NLP analysis engines, proved to be beneficial for clinical tasks. However, the drawback of these methods was the need to create numerous rules for each specific task, which made them less efficient.

Consequently, machine learning and deep learning-based approaches have gained significant interest in research for more efficient IE methods in the clinical domain \cite{wang2018clinical}. Support vector machine and conditional random field are widely used methods for detecting entities or events \cite{li2015end,rochefort2015accuracy}. Deep learning methods in the clinical domain have also benefited from transfer learning from the general domain. Apart from the popular BERT (Bidirectional Encoder Representations from Transformers) backbone\cite{devlin2018bert}, other strategies such as MT-clinical BERT that use multi-task learning for knowledge sharing among subtasks \cite{mulyar2021mt}, OIE4KGC that employs a UMLS knowledge-graph enhanced model \cite{muhammad2020open}, and self-supervised and retrieval-augmented methods have shown effectiveness \cite{chen2021disease}.

The applications of information extraction are diverse. For instance, phenotyping involves identifying diseases with specific features, like childhood obesity and neuropsychiatric disorders \cite{lyalina2013identifying}. IE can also be employed in feature engineering for the classification of cancer types and stages \cite{si2018frame,martinez2014cross}. Additionally, drug-related studies can draw on extracted examples for insights \cite{liu2013azdrugminer}, and the extracted information can be utilized to optimize clinical workflows or even provide clinical decision support \cite{gobeill2015exploiting,si2019enhancing}. 

However, as far as our knowledge extends, no existing methods have utilized QA to assist in EHR IE tasks, such as NER and relation extraction. The use of QA offers the advantage of simultaneously addressing all the subtasks in a more convenient manner.

\subsection{Question Answering}

In our focus on Machine Reading Comprehension (MRC) QA, the model provides an answer based on both the question and a given context. This QA task encompasses four main types of answers: yes-no questions, multi-choice questions, extractive questions, and generative questions \cite{jin2022biomedical}. For the first three types, models typically use only the encoder/embedding to obtain word representations. The widely adopted three-stage architecture proposed by FastQA \cite{weissenborn2017making} consists of semantic understanding, interaction, and prediction. Since the emergence of BERT-like models, enhancing the semantic understanding stage has become the mainstream approach to improving model performance \cite{liu2019roberta,lan2019albert,clark2020electra}. However, these three types differ in the predicting module. For instance, extractive QA predicts the probability of being start or end tokens, yes-no QA conducts binary classification \cite{clark2019boolq,yoon2019pre}, and multi-choice QA may select the most plausible hypothesis from constructed options \cite{li2021mlec,jin2020mmm}. The SQuAD 2.0 dataset introduced unanswerable questions for extractive QA \cite{rajpurkar2018know}, which is the task used to define our problem.

On the other hand, generative QA always requires an additional decoding module and is more commonly used in open-domain QA rather than machine reading comprehension. Interaction-aware embeddings are fed into a seq2seq decoder \cite{sutskever2014sequence}, transformer decoder \cite{vaswani2017attention}, or other structures like T5 \cite{raffel2020exploring}, GPT, etc. However, evaluating generative answers is challenging \cite{chen2019evaluating}, and it is even more difficult to determine whether the model provides unfounded answers. As a result, generative QA is not the preferred approach for our pipeline.

\section{Methods}

\begin{figure*}[h]
\centerline{\includegraphics[scale=0.4]{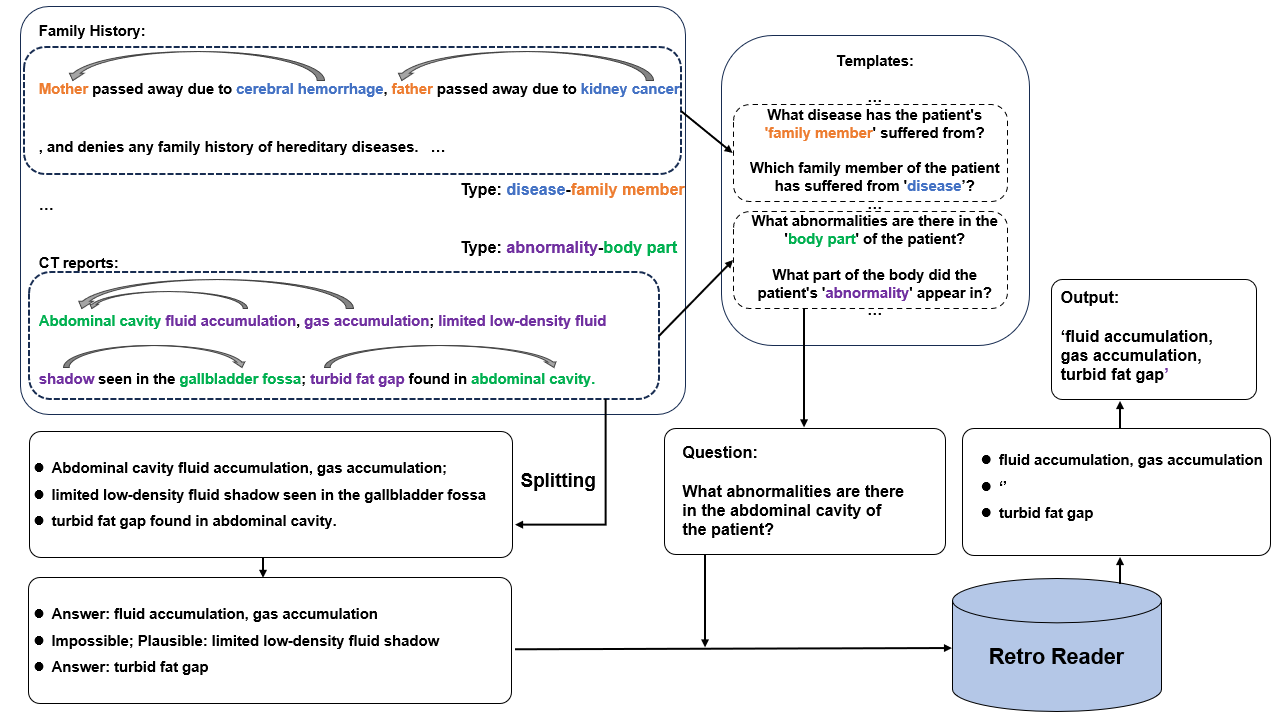}}
\caption{The architecture of our pipeline. From the EHR corpus, we obtain the original dependency annotations and the context. The words labeled by colors are the general entities annotated, each color stands for one type. The arrow represents the dependency annotations. During preprocessing, the dependency annotations are transformed into questions through manually constructed templates based on relation types. The contexts are split according to many-to-one correspondences of the relation pairs, resulting in sentence-level or paragraph-level texts. The questions and the texts are concatenated and sent into the QA model for training. We also introduce impossible questions with plausible answers through annotations of the same type. The QA model judges the answerability of each question-context pairs and output the answer span. Finally, the answers from split texts are merged to provide the final outputs.}
\label{fig2}
\end{figure*}

In this section, we will provide a detailed demonstration of the entire pipeline. The main purpose of the pipeline is to transform the dependency annotations into question-answer pairs, and help train a model for extracting information from the EHR data. The pipeline comprises three main components: 1. preprocessing of the dependency and textual data, 2. the QA model capable of discriminating unanswerable questions, 3. post-processing and combination of model outputs for application. The pipeline architecture and translated examples are displayed in Figure\ref{fig2}. For simplicity and privacy protection, we selected only several clauses from the original lengthy EHR paragraphs.

\subsection{Notation}

The raw annotations of our EHR data only compose of general entity annotations, which specify the textual content, type and start token of the entities, and the dependencies between them. The type of an entity $A$ is denoted as $t(A)$ and each dependency pair is denoted as $(A,B)$, indicating that a dependency relation exists from $A$ to $B$. Each sample pair $(A, B)$ is categorized into the class named "$t(A)$-$t(B)$". There are 15 entity types and 21 relation classes in total in the annotations.

Given a specific key (the information we are interested in), the output should consist of corresponding retrieved values. After preprocessing, the input should be the concatenation of a question $Q=(q_1,\cdots,q_n)$ together with a context $X=(X_1,\cdots,X_m)$. The pipeline output should include one or several spans $A_1=(X_{a_{1s}},\cdots,X_{a_{1e}}),\cdots, A_t=(X_{a_{ts}},\cdots,X_{a_{te}})$ from the context, where $a_{is}$ and $a_{ie}$ represent the start and end tokens of the $i^{th}$ span, respectively.

\subsection{Preprocessing}

The preprocessing consists of three parts which will be introduced below. We will still utilize the examples in Figure \ref{fig2}. 

\subsubsection{Transform dependency to question-answer pairs}

Firstly, we utilize pre-defined templates to convert the entity and dependency annotations into two kinds of question-answer pairs. Specifically, for disease, abnormality and body part entities, we can directly query them without specifying the dependency, for instance, "What abnormality does the patient have?". These questions equip the QA model with the ability to perform NER tasks effectively. For the dependency annotations, we manually design up to three question templates for each relation class "$t(A)$-$t(B)$" to query the left and right entities respectively. For example, in the first annotation in Figure \ref{fig2}, we can query the disease name by the template "What disease has the patient's 'family member' suffered from?', or reversely query the family member by  "Which family member of the patient has suffered from 'disease'?". This can capture bidirectional relationships and avoid annotation inconsistencies.

\subsubsection{Split paragraphs for discontinuous answer spans}

Secondly, in EHR, it is common to encounter scenarios where one abnormality corresponds to multiple body parts, or one body part may have multiple abnormalities. So the answer to the above question may have multiple spans. However, extractive QA, which deals with continuous spans, faces challenges in capturing discretely distributed answers. To address this issue, we handle cases of many-to-one correspondence during preprocessing as follows: If two answer spans are adjacent or separated only by punctuations, we merge them into one span: in Figure \ref{fig2} "fluid accumulation", "gas accumulation" are two annotated entities within the same body part, and they have merged to "fluid accumulation, gas accumulation". Additionally, when such correspondence still exists, we split the EHR paragraphs into sentences, and each sentence would have one or zero answer span to the question. The model, to be explained in the next subsection, is designed to provide an answer for answerable questions and an empty string for unanswerable ones. During evaluation, the output of each sentence will undergo post-processing to derive the final answer. So take the same example in Figure \ref{fig2}, the CT report is split into three sentences, where the first and third sentences both have one answer span to the question "What abnormalities are there in the abdominal cavity of the patient?" and the second sentence has zero. The final answer should be a combination of "fluid accumulation, gas accumulation" and "fat gap is turbid". The preprocessing strategy proves to be highly effective, as observed from the fact that, after merging adjacent spans, most of the dependency pairs follow a one-to-one correspondence within one sentence under EHR settings.

\subsubsection{Construct impossible questions}

Furthermore, to adhere to the format of the SQuAD dataset, we include "plausible answers" for impossible questions. For this purpose, we exclude the original impossible questions and sentences and instead construct new samples. Concretely, we select paragraphs containing more than one dependence pair of the same relation type. If the two left entities of the two pairs are different entities, we choose one question and consider the span of the answer to the other question as context. In the second example of Figure \ref{fig2}, "What abnormalities are there in the abdominal cavity of the patient?" is an impossible question to the clause "limited low-density fluid shadow seen in the gallbladder fossa;" but with a plausible answer "limited low-density fluid shadow" which has the same entity type "body part". The process forms an impossible question-answer pair with a seemingly plausible answer that differs from the original, thereby enhancing the model's capability to judge answerability.

After the preprocessing stage, we obtain question-answer pairs along with the context at the paragraph or sentence level. The answer can either be empty or represented by a continuous span. These processed data are then fed into the model. 

\subsection{QA Model}

\begin{figure}[ht]
\centerline{\includegraphics[scale=0.55]{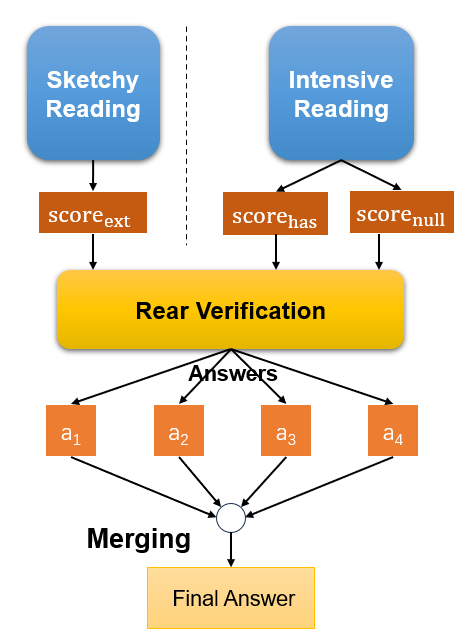}}
\caption{The Retro-reader model contains two reading modules and a rear verification module. The scores produced by reading modules will be compared to a threshold to decide whether the question is answerable.}
\label{fig1}
\end{figure}

The main body of the Question Answering (QA) model we employ is the original Retro-Reader\cite{zhang2021retrospective}, one of the benchmark models for SQuAD 2.0 (Stanford Question Answering Dataset). And we will give a brief introduction to the model. Retro-Reader comprises two primary components: a sketch reading module responsible for making a coarse judgment about answerability, and an intensive reading module that generates candidate answer spans. Additionally, it has integrated two threshold-based verification modules to further enhance the model's capability for predicting answerability. The architecture can be concluded in Figure \ref{fig1}.

In the sketchy reading module, the input is first encoded by a pretrained multi-layer Transformer\cite{vaswani2017attention,devlin2018bert}. The last layer of hidden states $\bm{H}_1$ is used as the embedding, which is sent to a fully connected layer for a two-way classification. And the predicted logits of not having an answer $\hat{y}_i$ are used for loss computation, where $y_i\in\{0,1\}$ is the true label:
\begin{equation}
    L_{ans1} = -\frac1N\sum_{i=1}^N(y_i\log \hat{y}_i + (1-y_i\log(1-\hat{y}_i))
\end{equation}

The intensive reading module uses another Transformer but starts from the same pretrained weights. We denote the embeddings by $\bm{H}_2 = (H_Q,H_P)$, standing for the question and passage respectively. The module adopts cross attention to compute the question-aware representation of the texts, which set $Q=H, K=V=H_Q$ in the multi-head attention in \cite{vaswani2017attention}. The final embedding $H'$ after being weighted by attention is applied to two linear layers with a softmax operator to compute start and end position probabilities. The loss of the span prediction is written as:
\begin{equation}
    L_{span} = -\frac1N\sum_{i=1}^N[\log(s_{y_i^s})+\log(e_{y_i^e})]
\end{equation}

which is the sum of the log probabilities of the true start and end positions. Another binary cross entropy loss $L_{ans2}$ is introduced to render this module also able to identify unanswerable questions. So the loss of this whole module is:
\begin{equation}
    L = \alpha_1L_{span}+\alpha_2L_{ans2}
\end{equation}

For our usage, both reading modules are trained on sentence-level and paragraph-level data to deal with multi-granularity in preprocessed data. As the answerability judgment is essential in the pipeline, we slightly enlarge the weight of $L_{ans2}$. Finally, three scores: an external verification score, a has-answer score and a no-answer score are computed for rear verification:
\begin{equation}\begin{split}
    score_{ext} & = \hat{y}_i - (1-\hat{y}_i)\\ 
    score_{has} & = \max (s_k + e_l),\, 1<k<l\le n \\
    score_{null} & = s_1 + e_1
\end{split}\end{equation}

Here $s_1,e_1$ is the start and end possibility of the [CLS] token. If the difference score $score_{diff} = score_{null}-score_{has}$ and the mixture score $\beta_1score_{diff} + \beta_2score_{ext} $ are both larger than a chosen threshold $\delta$, the QA model will output the answer span. Otherwise, it will predict the null string.

\subsection{Postprocessing and Application}

To ensure consistency in preprocessing and evaluation, we establish the format of the gold standard. The answer should be a concatenation of the initially annotated general entities, with one comma placed between each entity. The presence of the comma will be taken into account when computing the EM (exact match) score but not in the computation of the F1 score. If the answer of one sentence is empty, it will not be concatenated. In the case where all the answers of the sentences are empty, we consider the question to be unanswerable.

In a real-world setting, conducting comprehensive information extraction for all the EHRs is essential. Depending on the type of EHR, different templates will be applied and queried. If the EHR contains disease, abnormality and body part information, the NER step should be performed first, and the resulting entities will be utilized in the templates to query the text. Otherwise, the template itself with several fed-in words is sufficient for the query. Once all the possible queries are listed, we split the paragraph, obtain the model outputs, and merge them to complete the task. Questions that yield an empty string as the final answer are dismissed as non-informative.

\section{Experiments}

\subsection{Data and Annotations}

We utilized real de-identified EHR data from a single hospital as the basis for our study. The corpus comprises various EHR types, such as past medical history, personal history, family history, CT reports, and radiology reports. Our annotators were instructed to first label the general entities and then establish the dependency relationships between them. The identified entities encompass 15 types, including numbers, sizes, trends, properties of abnormalities, diseases, body parts, immune groups, and values. These general entities are further connected to form 21 distinct relation classes. For training and testing purposes, we selected 18 high-frequency classes.

Following preprocessing steps, we compiled a dataset consisting of 1718 EHR passages, resulting in a total of 14528 question-answer pairs based on dependencies. The training set consists of 11451 QA pairs, while the dev set contains 1510 pairs, and the test set contains 1567 pairs. Notably, the dataset contains 3576 impossible questions, with 350 instances each in the dev and test sets. Additionally, the dataset includes 5600 NER-like questions found in 908 passages. To handle the two types of questions we process them separately.

\subsection{Implementation}

The model backbone is detailed in Section III. For recognition of Chinese EHR texts, we employed the "roberta-base-chinese" and "bert-base-chinese" pretrained embeddings. To optimize performance, the model was deployed in parallel on two GTX 2080Ti 11G GPUs. Each GPU was assigned a train and eval batch size of 2, and the max length was set to 512. The training process encompassed the entire training set for four epochs. The training was completed within approximately 40 minutes, while producing predictions for evaluation took around five minutes. We set $\alpha_2=0.8$ in the intensive reading module loss, and all other hyperparameters remained consistent with those utilized in \cite{zhang2021retrospective}.

\section{Results}

\subsection{NER QA Results}

The NER-like QA evaluation is tested only on questions that pertain to diseases, abnormalities and body parts, the only three entity types not categorized into general entities or events. The evaluation standard we adopt involves both entity-level Exact Match (EM) score and token-level micro F1 score. A token is considered a True Positive if both its position and label are correct, and a False Negative if either of them mismatches.

For Chinese NER, we opted to use the most commonly employed baseline model, which consists of BERT/RoBERTa (Chinese) in combination with a classifier (linear layer) for NER sequence tagging. The performance of each baseline model as well as our pipeline is summarized in Table \ref{tab2}.

\begin{table}[htbp]
\linespread{1.3}\selectfont
\caption{NER results comparison}
\begin{center}
\begin{tabular}{|c|c|c|}
\hline
\multicolumn{1}{|c|}{\multirow{2}{*}{\textbf{Model Names}}}&\multicolumn{2}{|c|}{\textbf{Results}} \\
\cline{2-3} 
 & EM& F1 \\
\hline
BERT-base-chinese + linear& 0.906 & 0.815   \\
\hline
RoBERTa-base-chinese + linear& 0.851 & 0.933  \\
\hline
RoBERTa-base-QA pipeline w/o finetuning& 0.166 & 0.255  \\
\hline
RoBERTa-base-QA pipeline (ours)& \textbf{0.894} & \textbf{0.965}  \\
\hline
\multicolumn{3}{l}{$^{\mathrm{a}}$ The F1 and EM score of the baseline models and our pipeline.} \\
\end{tabular}
\label{tab2}
\end{center}
\end{table}

The initial performance of the QA pipeline is notably poor when applied without finetuning. This discrepancy in results can be attributed to the inherent differences between the clinical domain and the general domain. However, following the training process, our pipeline demonstrates significant improvement and achieves competitive results on the NER task. 

\subsection{Relation-like QA Results}

In this section, we evaluate the other question types for information extraction collectively. For evaluation, we employ the F1 and EM metrics. The token-level F1 computation is similar to the NER task, with the distinction that QA does not involve entity class labels. Since the extraction of information within given keys cannot be directly classified into existing tasks, such as Open IE or relation extraction, we establish our baseline model as a pipeline without any training.

\begin{table}[htbp]
\linespread{1.3}\selectfont
\caption{Relation QA results comparison}
\begin{center}
\begin{tabular}{|c|c|c|}
\hline
\multicolumn{1}{|c|}{\multirow{2}{*}{\textbf{Model Names}}}&\multicolumn{2}{|c|}{\textbf{Results}} \\
\cline{2-3} 
 & EM& F1 \\
\hline
RoBERTa-base-QA pipeline w/o finetuning& 0.363 & 0.514  \\
\hline
RoBERTa-base-QA pipeline (ours)& \textbf{0.922} & \textbf{0.953}  \\
\hline 
\end{tabular}
\label{tab3}
\end{center}
\end{table}

\begin{table}[htbp]
\linespread{1.3}\selectfont
\caption{Answerability judgment results comparison}
\begin{center}
\begin{tabular}{|c|c|}
\hline
\textbf{Model Names}&\textbf{Accuracy} \\
\hline
RoBERTa-base-QA pipeline w/o finetuning& 0.919   \\
\hline
RoBERTa-base-QA pipeline (ours)& \textbf{0.998}   \\
\hline 
\end{tabular}
\label{tab4}
\end{center}
\end{table}

\begin{figure}[htbp]
\centerline{\includegraphics[scale=0.21]{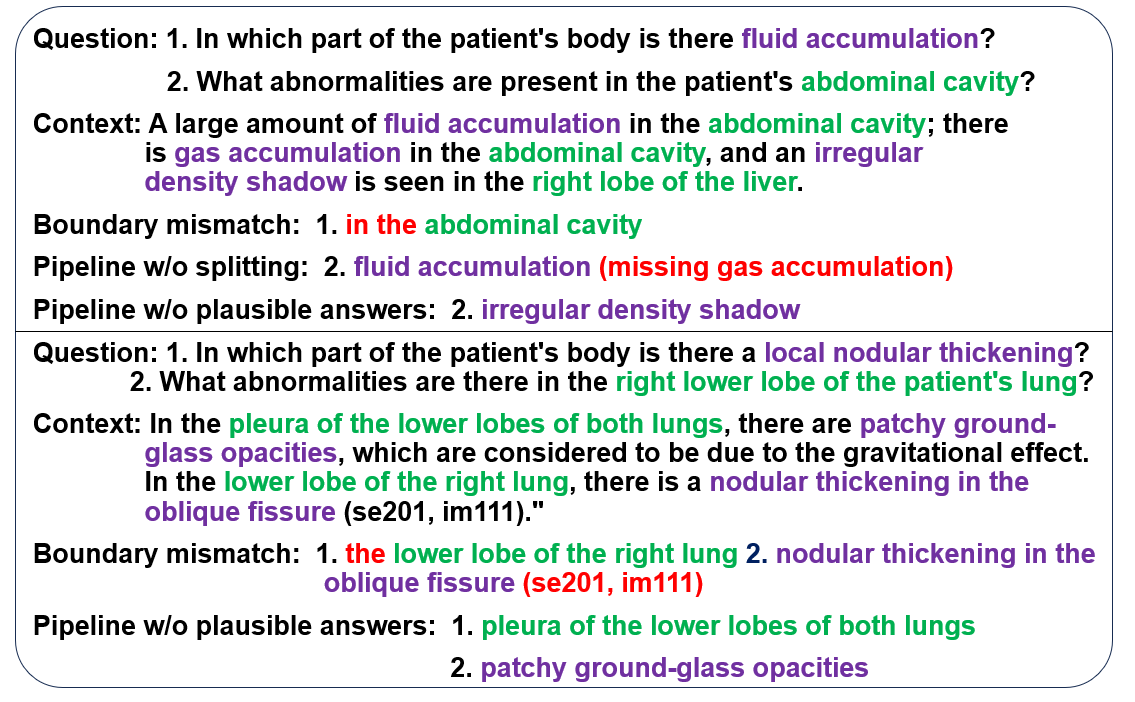}}
\caption{The examples for translated gold annotations and predictions of different models. The colored words in the context are the gold answer, and the red word is where the boundary mismatches. Note that translation may omit some expressions in Chinese, so the examples are for reference only and may not reflect the whole picture of Chinese EHR texts.}
\label{fig3}
\end{figure}

The performance improvement in Table \ref{tab3} demonstrates the effectiveness of the processing pipeline, which render the model able to answer EHR-related questions by constructing such question-answer pairs from original data. We also manually checked the model's predictions. Among the 122 samples in the test set that do not exactly match the gold standard, approximately 2/3 of the errors result from a margin mismatch of one to two tokens. These errors also include some subtle inconsistencies among the annotations. We provide some examples in Figure \ref{fig3}, where the "Boundary Mismatch" line illustrates these cases. The figure caption contains the translations. The red-labeled characters representing the mismatch may be locative prepositions or function words. Both predicted spans are plausible in these cases. So intuitively, the performance seems even better than indicated and appears suitable for practical usage. Other main causes of error are finding the wrong correspondence to another left entity of the same entity type at the paragraph level, and predicting a long sequence unexpectedly while the true answer is short. The accuracy of judgment, on the other hand, turned out to be very close to 1 after finetuning as shown in Table \ref{tab4}. This actually guarantees the validity of merging sentence-level outputs.

\subsection{Ablation Study}

n this section, we aim to validate the necessity of our design, which consists of two main components in the preprocessing stage. The first component involves splitting some paragraphs into sentences, which addresses many-to-one correspondence and discontinuous answer spans. The second component is the construction of impossible questions with plausible answers, which enhances the model's ability to capture dependencies and discriminate answerability.

Table \ref{tab5} and Table \ref{tab9} provide evidence of the design's effectiveness. Each component contributes to a gain of approximately $0.02$ to $0.05$ in the EM score and $0.05$ to $0.08$ in the F1 score for relation-like questions. However, for NER-like questions, the performance experiences a significant drop without the component. This drop can be attributed to the fact that most paragraphs contain more than five entities of the same type, making it challenging for the model to handle NER-like questions effectively. It's worth noting that the ablation of plausible answers is not conducted for NER-like questions due to the absence of plausible answers when a specific entity type does not exist in the sentence.

\begin{table}[htbp]
\linespread{1.3}\selectfont
\caption{Ablation results comparison on relation-like questions}
\begin{center}
\begin{tabular}{|c|c|c|}
\hline
\multicolumn{1}{|c|}{\multirow{2}{*}{\textbf{Model Names}}}&\multicolumn{2}{|c|}{\textbf{Results}} \\
\cline{2-3} 
 & EM& F1 \\
\hline
RoBERTa-base-QA pipeline w/o splitting& 0.844 & 0.88   \\
\hline
RoBERTa-base-QA pipeline w/o plausible answers& 0.88 & 0.921  \\
\hline
RoBERTa-base-QA pipeline (ours)& \textbf{0.922} & \textbf{0.953}  \\
\hline
\end{tabular}
\label{tab5}
\end{center}
\end{table}

\begin{table}[htbp]
\linespread{1.3}\selectfont
\caption{Ablation results comparison on NER questions}
\begin{center}
\begin{tabular}{|c|c|c|}
\hline
\multicolumn{1}{|c|}{\multirow{2}{*}{\textbf{Model Names}}}&\multicolumn{2}{|c|}{\textbf{Results}} \\
\cline{2-3} 
 & EM& F1 \\
\hline
RoBERTa-base-QA pipeline w/o splitting& 0.189 & 0.322   \\
\hline
RoBERTa-base-QA pipeline w/o plausible answers& - & -  \\
\hline
RoBERTa-base-QA pipeline (ours)& \textbf{0.894} & \textbf{0.965}  \\
\hline
\multicolumn{3}{l}{$^{\mathrm{a}}$ Here '-' means the ablation is not carried out} \\
\end{tabular}
\label{tab9}
\end{center}
\end{table}

A case study can help us better understand how these tricks work. In Figure \ref{fig3}, we present two instances to illustrate their effectiveness. In the first example, we have two abnormalities that point to the same body location but are mentioned in two separate clauses. If we didn't perform the splitting operation during preprocessing, the model would likely predict only one of the abnormalities. We observed similar cases even when the two spans are adjacent in the original annotations, as they are initially annotated in this way. 

In both examples, there are two clauses in analog structures. When the model lacks fine-tuning on impossible questions with plausible answers, there is a higher probability of mixing up the relativity between the clauses and predicting the wrong span of the same type. So both of the tricks are necessary to achieve more stable and accurate extractions.

\subsection{Generalization Performance}

Finally, we aim to explore whether the model can be utilized to extract information beyond the annotations, simulating a zero-shot setting. For this purpose, we manually propose 40 questions not covered in the 18 relation classes. These questions comprise 20 wh-questions and 20 yes-no questions. As our pipeline relies on extractive Question-Answering (QA), we expect the model to extract the negation word if the answer is "no," and extract the relevant statement if the answer is "yes." Therefore, if the model predicts a negation word, we consider its answer as "no," and as "yes" otherwise.

\begin{table}[htbp]
\linespread{1.3}\selectfont
\caption{Genralization results}
\begin{center}
\begin{tabular}{|c|c|c|}
\hline
\multicolumn{1}{|c|}{\multirow{2}{*}{\textbf{Tasks}}}&\multicolumn{2}{|c|}{\textbf{Results}} \\
\cline{2-3} 
 & EM& F1 \\
\hline
zero-shot questions& 0.65 & 0.763   \\
\hline
yes-no questions& 0.85 & -  \\
\hline
\multicolumn{3}{l}{$^{\mathrm{a}}$ Here '-' means not able to compute the F1 score.} \\
\end{tabular}
\label{tab6}
\end{center}
\end{table}

The generalization results are presented in Table \ref{tab6}. The model performs well on yes-no questions, achieving satisfactory accuracy. However, it only performs acceptably on other types of questions. The primary reason for this discrepancy is that the model lacks an understanding of the EHR writing format during the fine-tuning process, making it challenging for it to answer questions related to overall information.

\subsection{Discussion and Limitations}

The results obtained from our pipeline demonstrate its superiority, meeting most of the requirements in our task settings. However, due to limitations in annotations and equipment, several potential improvements are not explored.

Firstly, we acknowledge the need to address a combination of yes-no questions, multiple-choice questions, and extractive questions within the context of EHR extraction. In certain situations, while extracting information from EHRs, we may encounter queries related to the patient's diagnosis of specific diseases, and disease classification or severity, which typically require constrained options for answers. Although our pipeline showcased competence in handling yes-no questions, it would be better to apply targeted models. An additional module, such as rule-based or BERT-based classifiers, could achieve high accuracy in distinguishing among these three question classes. By incorporating such classifiers, the models described in Section II can be adapted accordingly to form again a unified framework.

Secondly, we recognize the importance of enabling few-shot or even zero-shot capabilities for our models. In real-world scenarios, medical concepts may lack available annotated data\cite{sivarajkumar2022healthprompt}, or the existing data may not align precisely with practical requirements.  Pretrained Language models possess the transferring function, and LLMs even have emergent ablities\cite{wei2022emergent}. Prompts or other tricks may be invented to trigger such capacity. As new techniques continue to emerge, other avenues for improvement may become possible, ultimately leading to the development of a more comprehensive and powerful EHR IE system in the foreseeable future.

\section{Conclusion}

In this paper, we present a pipeline for comprehensive information extraction from Electronic Health Records using question-answering models. Our approach involves three main steps: paragraph splitting, impossible question construction, and output merging. This pipeline is designed to handle unified Named Entity Recognition and dependency-based extraction, and effectively address the issue of discontinuous multi-span answers. By fine-tuning the pipeline on our Electronic Health Records data, we achieve competitive performance on both types of questions, demonstrating a certain level of generalization to answer zero-shot and yes-no questions. Consequently, the pipeline proves to be practically useful for real-world tasks.

\medskip

\bibliographystyle{IEEEtran}
\bibliography{bib}

\end{document}